\documentclass{article}

\usepackage[preprint, nonatbib]{neurips_2023}

\usepackage[utf8]{inputenc} 
\usepackage[T1]{fontenc}    
\usepackage{hyperref}       
\usepackage{url}            
\usepackage{booktabs}       
\usepackage{amsfonts}       
\usepackage{nicefrac}       
\usepackage{microtype}      
\usepackage{xcolor}         
\usepackage{amsmath}
\usepackage[ruled,vlined]{algorithm2e}
\usepackage{graphicx}
\usepackage{enumitem} 
\usepackage{amsthm}

\newtheorem{theorem}{Theorem} 
\newtheorem{lemma}{Lemma}     
\newtheorem{assumption}{Assumption} 

\title{FedStaleWeight: Buffered Asynchronous Federated Learning with Fair Aggregation via Staleness Reweighting}

\DeclareMathOperator*{\argmin}{arg\,min}

\author{%
  Jeffrey Ma \quad Alan Tu \quad Yiling Chen \quad Vijay Janapa Reddi\\
  Harvard University\\
  Cambridge, MA 02134 \\
  \texttt{jeffreyma@g.harvard.edu, alantu@college.harvard.edu} \\
  \texttt{yiling@seas.harvard.edu, vj@eecs.harvard.edu}
}

\begin{document}

\maketitle

\begin{abstract}
Federated Learning (FL) endeavors to harness decentralized data while preserving privacy, facing challenges of performance, scalability, and collaboration. Asynchronous Federated Learning (AFL) methods have emerged as promising alternatives to their synchronous counterparts bounded by the slowest agent, yet they add additional challenges in convergence guarantees, fairness with respect to compute heterogeneity, and incorporation of staleness in aggregated updates. Specifically, AFL biases model training heavily towards agents who can produce updates faster, leaving slower agents behind, who often also have differently distributed data which is not learned by the global model. Naively upweighting introduces incentive issues, where true fast updating agents may falsely report updates at a slower speed to increase their contribution to model training. We introduce FedStaleWeight, an algorithm addressing fairness in aggregating asynchronous client updates by employing average staleness to compute fair re-weightings. FedStaleWeight reframes asynchronous federated learning aggregation as a mechanism design problem, devising a weighting strategy that incentivizes truthful compute speed reporting without favoring faster update-producing agents by upweighting agent updates based on staleness. Leveraging only observed agent update staleness, FedStaleWeight results in more equitable aggregation on a per-agent basis. We both provide theoretical convergence guarantees in the smooth, non-convex setting and empirically compare FedStaleWeight against the commonly used asynchronous FedBuff with gradient averaging, demonstrating how it achieves stronger fairness, expediting convergence to a higher global model accuracy. Finally, we provide an open-source test bench to facilitate exploration of buffered AFL aggregation strategies, fostering further research in asynchronous federated learning paradigms.\footnote{Code for experiments can be found at \href{https://github.com/18jeffreyma/afl-bench}{https://github.com/18jeffreyma/afl-bench}}
\end{abstract}

\section{Introduction}

\subsection{Motivation and Background}

In real-world scenarios, the task of training a model may be distributed among up to millions of agents, called edge devices, each of whom contribute their own limited (and perhaps private) set of data, a paradigm known as federated learning (FL). Our paper focuses on cross-device FL, in which there must be a large number of agents in order to extract meaningful results. The other type of FL, cross-silo, utilizes fewer agents because each one has more data and participates actively in the model training process, like a consortium of hospitals sharing information about a new disease \cite{huang2022crosssilo}.

Cross-device FL faces two major challenges when put into practice: scalability and privacy. In a massive distributed network, it's impossible for every agent to be active at every time step, or for all agents to train their private models at the same speed. Naive aggregation techniques rely on concurrent reports from agents because, for instance, a simple average may be taken over reports in order to update the global model at regular intervals. By relaxing the concurrency constraint, we move away from synchronous FL, and the problem becomes how to best aggregate updates arriving at different points in time. On the topic of privacy, agents want their personal data to remain fully confidential; the global model must therefore evolve in a way that does not reveal individual contributions through gradient updates or the like \cite{nguyen2022federated}. Recent work in secure aggregation (SecAgg) and differential privacy provide frameworks for exploring the tradeoffs between privacy and utility \cite{kadhe2020fastsecagg}.

Unlike its synchronous conterpart, Asynchronous federated learning (AFL) does not rely on clients to all finish and communicate their results in the same round and differs from its synchronous counterpart in the following ways:
\begin{itemize}[leftmargin=1em]
\item\emph{Synchronicity:} Clients are assumed to be heterogeneous and may return updates at different times or fail to return updates in time, necessitating mechanisms that do not wait for all participants before updating the global model through aggregation.
\item\emph{Staleness:} As models may take longer to update, AFL mechanisms often are able to merge updates to models (i.e. model version $T$ is deployed to a client, the client takes a long time to fit, and it replies with its updated model while the global server has already aggregated updates and stepped to model version $T+1$). Emerging research in AFL hypothesizes that these stale updates are likely still useful and should be incorporated into learning.
\item\emph{Convergence challenges:} AFL often experiences higher training instability based on the strategy chosen for aggregating stale updates, similar to issues encountered in off-policy reinforcement learning (RL).
\end{itemize}

\subsection{Our Contribution}

We observe that the asynchronicity in AFL naturally leaves slow updating agents behind: since updates are no longer incorporated synchronously and are incorporated on demand, agents who are not able to update as quickly will contribute less to global model learning. However, naively up-weighting slow agent updates creates incentive issues: if our upweighting is too extreme, we incentivize fast updating agents to throttle and falsely report at slower speeds to increase their contribution to learning.

Thus, we explore the problem of fairly aggregating updates from clients in AFL to offset the unfairness introduced by compute heterogeneity in asynchronous federated learning. Several known synchronous approaches such as FedAvg and FedProx exist and more nuanced asynchronous approaches have been explored such as a fixed buffer length with averaging \cite{nguyen2022federated} or simply updating the central model in an asynchronous immediate manner \cite{chen2020asynchronous}. To our knowledge, more nuanced weighting of stale updates or an adaptive approach has not yet been explored. Our contribution is the following:

\begin{enumerate}[leftmargin=2em]
    \item We reframe the problem of \emph{asynchronous federated learning aggregation} as a welfare maximization problem and derive a gradient aggregation method which is strategy-proof to fast agents who may throttle their update rate to have their updates weighted higher while increasing the contribution of slower update producing agents to global model training.
    \item We show how our derived fair weighting can be computed \emph{via observing agent update staleness}, a quantity naturally visible in buffered asynchronous federated learning, and present an algorithm for fair federated learning, FedStaleWeight.
    \item We provide a convergence guaruntee for FedStaleWeight in the smooth, non-convex setting, showing that the upweighting of stale model updates can still yield convergence.
    \item We empirically show in a series of realistic non-IID federated learning settings how current AFL aggregation techniques are \emph{biased towards clients with higher throughput} and how FedStaleWeight maintains fairness and subsequently \emph{converges to higher global model accuracy faster}. We release an open-source test bench for other researchers to explore further buffered AFL aggregation strategies.
\end{enumerate}

\section{Related Work}

{\bf Synchronous FL.} Synchronous FL excels in the privacy dimension: when the principal must wait for all agents before aggregating their data, it becomes nearly impossible to recover an individual contribution from an average over so many points. The slowest agent, however, determines the training pace of the global model, resulting in a bottleneck. If the principal were to perform aggregation with only, say, the fastest half of agents, then an element of selection bias is introduced. These concerns aside, numerous synchronous techniques have found experimental success, and their primary features can be summarized as follows:

\begin{itemize}[leftmargin=1em]
    \item \emph{FedAvg: }McMahan et al. (2016) coined the term ``federated learning'' and suggest a simple weighted average of gradients to update the global model, where the weights correspond to the $p_i$ terms in the overall objective described in the next section \cite{mcmahan2023communicationefficient}. 
    \item \emph{FedProx: }Li et al. (2020) improve upon FedAvg by adding a regularization term $\frac{\mu}{2}\|\bf{x} - \bf{x_g}\|^2$ to each local objective $F_i$, where $\bf{x_g}$ is the current global model. This allows each local model to more resemble the global model but converges under a more restrictive set of assumptions \cite{li2020federated}.
    \item \emph{FedAvgM: }Hsu et al. (2019) also employ a form of regularization through server momentum, which accumulates past gradients in order to stabilize and accelerate training \cite{hsu2019measuring}. 
    \item \emph{FedNova: }Wang et al. (2020) note that heterogeneous update speeds can cause objective inconsistency, where the server model approaches a local optimum that does not match the global optimum of the original objective function \cite{wang2020tackling}. The authors provide a formal analysis of this phenomenon and propose FedNova, which normalizes each client's gradient by the number of updates that the client has sent in the current round before taking the overall average for a round. This technique encompasses both FedAvg and FedProx and has been shown to outperform both of them.
\end{itemize}

{\bf Asynchronous FL.} AFL is a natural fit in the cross-device setting due to heterogeneity among agents' availability as well as training and data transfer speeds. Here, we provide an overview of the most prominent work in AFL to date:

\begin{itemize}[leftmargin=1em]
    \item \emph{FedAsync: }Xie et al. (2019) gave the first convergence guarantees for the AFL problem; their \emph{fully} asynchronous method forces the server to update its model after every client update \cite{xie2019asynchronous}. As later works observe, this is not only computationally taxing but insecure.
    \item \emph{ASO-Fed: }Chen et al. (2020) deal with the special case of online learning, where each agent's local dataset is continuously growing as new observations arrive \cite{chen2020asynchronous}. Newer data need not stem from the same distribution as older data, which means that a robust AFL framework must be able to handle non-IID data.
    \item \emph{FedAdaGrad, FedAdam, FedYogi: }Reddi et al. (2020) tailor the well-known adaptive optimizers AdaGrad, Adam, and Yogi to the FL problem \cite{reddi2020adaptive}. 
    \item \emph{SAFA: }Wu et al. (2020) propose Semi-Asynchronous Federated Averaging (SAFA) to deal with the issues of stragglers, dropouts, and staleness \cite{wu2020safa}.
    \item \emph{Pisces: }Jiang et al. (2022) use a custom scoring mechanism to choose which agents will participate in the global model training at a given time step \cite{jiang2022pisces}. Their utility function attempts to distinguish between slow clients with high-quality data and those which are just slow. 
    \item \emph{AsyncFedED: }Wang et al. (2022) aggregate client data by calculating a ``Euclidean distance'' between stale models and the global model and then weighting each client's update appropriately \cite{wang2022asyncfeded}.
    \item \emph{FedBuff: }In order to avoid full asynchronicity, Nguyen et al. (2022) hold agent updates in a buffer. Only when the buffer is full does the data get aggregated and pushed to the server \cite{nguyen2022federated}. This work directly inspired and most closely resembles our work. SecAgg \cite{bonawitz2016practical} \cite{guo2021secure} can also be integrated to maintain privacy guarantees. 
    \item \emph{FedFix: }Fraboni et al. (2023) try to unify a large number of the above methods through their FedFix framework: it uses stochastic weights during aggregation \cite{fraboni2023general}.
\end{itemize}

\section{Preliminaries}

\subsection{Asynchronous FL Setting}

The federated optimization problem with $m$ agents can be formulated as:
\begin{align}
\min_{\mathbf{x}\in\mathbb{R}^d} \left[f(\mathbf{x}) := \sum_{i=1}^m p_i F_i(\mathbf{x})\right]
\end{align}
where $p_i$ is the importance of the $i$th agent (often assumed to be in proportion to the size of each client's dataset, i.e. $p_i = n_i/n$ or equal for all agents $p_i = 1/n$). $F_i$ represents agent $i$'s local objective function and is only available to agent $i$.

\begin{figure}[htbp]
    \centering
    \includegraphics[width=0.98\textwidth]{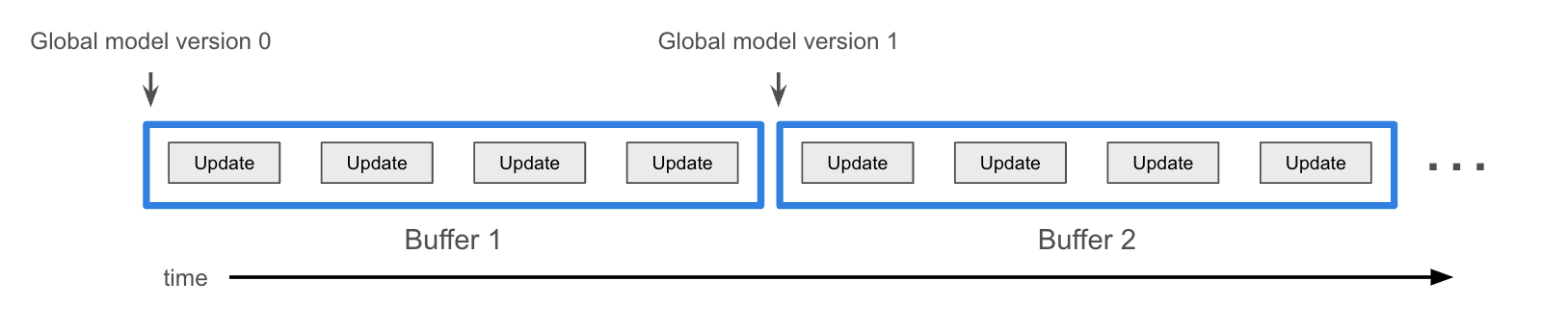}
    \caption{Aggregation in buffered asynchronous federated learning. Updates accumulate in the buffer over time: for a buffer size of $b=4$, the diagram shows aggregation of updates and how model versions are incremented for each arriving update.}
    \label{fig:buffer_diagram}
\end{figure}

Specifically, we study the buffered asynchronous federated learning setting, consisting of a singular central \emph{server} with an update aggregation buffer and $m$ clients. The server only updates once $b$ updates arrive at the buffer, where $b$ is a tune-able parameter. \emph{Each client} has their own private data and pulls the latest version of the global model, trains it on its private dataset and communicates the update back to the server, which is appended to the buffer to later be aggregated (Figure \ref{fig:buffer_diagram}). Specifically, in the asynchronous setting, we introduce the challenge of \emph{staleness}, where updates to update model at version $v$ may have been trained on models of version smaller than $v$. 

In the buffered AFL setting, we denote \emph{staleness} $\tau_i$ for an agent $i$'s update as the positive difference in version number between the current global model and the global model that the update was computed from. Formally, we denote a global model with version $n$ and with 0 client training steps as $w_{n, 0}$. For a client $i$, we denote the global model with version $n'$ before and after training on $q$ local steps as $w_{n', 0}$ and $w_{n', q}$, respectively; the client's update is denoted as $\Delta_i = w_{n', q} - w_{n', 0}$. Given the current global model $w_{n, 0}$ and update $\Delta_i$, the staleness is thus $\tau_i = n - n'$.

\subsection{Defining Fairness in Update Aggregation} \label{sec:fairness}


Each FL client $i$ represents its true update reporting speed as a random variable $\mathbf{R_i}$ where we denote the mean reporting speed as $r_i = \mathbb{E}[\mathbf{R_i}]$. Each agent $i$ chooses to report a random variable $\tilde{\mathbf{R_i}}$ with mean $\tilde{r_i} = \mathbb{E}[\tilde{\mathbf{R_i}}]$ as their observed update reporting speed. We assume for simplicity that $\tilde{r_i} \in [0, r_i]$, following intuition of fixed hardware specs: in other words, clients cannot manipulate to be faster on average than their fastest (true) compute speed on average. Given reported update speeds, we naturally see that the expected proportion that each agent contributes to learning without re-weighting is:
\begin{equation}
    p_i(\tilde{r_i}) = \frac{\tilde{r_i}}{\sum_{i=1}^n{\tilde{r_i}}}
\end{equation}

A fair re-weighting $\alpha(\cdot)$ takes $p_i(\tilde{r_i})$ and returns a weighting close to equal for every participating learning agent. In other words, we seek an $\alpha(\cdot)$ that solves the following constrained welfare maximization problem:
\begin{equation}
    \min_{\alpha(\cdot)} \sum_{i=1}^n{\left\lVert \alpha(p_i(\tilde{r_i})) p_i(\tilde{r_i}) - \frac{1}{n} \right\rVert_2} \quad \text{ such that } \quad \sum_{i=1}^n{\alpha(p_i(\tilde{r_i}))p(\tilde{r_i})} = 1
\end{equation}
Clients seek to maximize a utility representing their effective influence (and thus their accuracy), modelled as their re-weighting times the overall proportional influence of their reported mean update rate. We use this utility function to justify strategy-proofness in Section \ref{sec:online-learning}.
\begin{equation}
    \max_{\tilde{r_i}  \in [0, r_i]}{\alpha(p_i(\tilde{r_i})) p_i(\tilde{r_i})}
\end{equation}

\subsection{Step Size as Uncertainty in Approximation} \label{sec:step-size}
Recall that computing the update step $u^*$ for a client with loss $L(\mathbf{x})$ at point $\mathbf{x}$ can be framed as the solution of the following minimization game, where step size $\eta$ represents the uncertainty in using the approximation $L(\mathbf{x} + u^*) \approx L(\mathbf{x}) + u^{*, \top}\nabla L(\mathbf{x})$ for stochastic gradient descent:
\begin{gather}
    u^* =\argmin_{u \in \mathbb{R}^{d}}\left\{ L(x) + u^{\top} \nabla L(x) + \frac{1}{2\eta} u^{\top} u\right\}
\end{gather}
Similarly, in federated learning, we wish to restrict our aggregated gradient to be no larger than the step sizes of any of the individual aggregated gradients for convergence reasons, necessitating that any solution $\alpha(\cdot)$ must have $\sum_{i=1}^n{\alpha(p_i(\tilde{r_i}))p(\tilde{r_i})} = 1$ as noted above.

\section{Algorithm}

\subsection{Solving the Welfare Maximization Problem}

Naively, if $\alpha(\cdot) = 1$ is chosen (all agents are weighted equally), we see that $p(\tilde{r_i})$ is strictly increasing with respect to $\tilde{r_i}$ and all agents will report $\tilde{r_i} = r_i$. However, we see that welfare is poor, since an agent's influence $\alpha(p(\tilde{r_i}))p(\tilde{r_i})$ is exactly increasing with respect to its rate, meaning that slow agents are left behind as they cannot manipulate any higher than their true $r_i$.

We observe that $\alpha(\cdot)$ should be monotonically decreasing with respect to its input, but not at a rate that decreases slower than $p(\cdot)$ increases. If the latter as true, faster agents would be incentivized to throttle their rates to improve their own weighted influence. The optimal $\alpha(\cdot)$ becomes $\alpha(\cdot) = \frac{1}{n(\cdot)}$, where $\alpha(p_i(\tilde{r_i})) = 1/(np_i(\tilde{r_i}))$. Thus, with this choice, each agent's weighted influence remains constant regardless of which reported $\tilde{\mathbf{R_i}}$ they choose, and thus there is no incentive to not report truthfully. We note that this weighting achieves \emph{weak truthfulness} where agents are indifferent between reporting at any speed, since their utility stays the same regardless.

\subsection{Deriving Expected Influence From Staleness}

In buffered AFL, we assume the global aggregator has access only to the aggregation buffer and not the arrivals of individual agent updates, disallowing use of a weighting scheme that relies on estimating individual agent reporting schemes based on unique identifiers. We proceed to show how a fair weighting can be computed from staleness, an available local quantity, in asynchronous federated learning.  

Given an initially empty buffer, we can compute the expected index of the first update from a given agent $i$ with mean rate $\tilde{r_i}$ by summing the expected number of updates from all agents in the expected time period for one update from agent $i$. For example, we note that an agent twice as fast as $i$ would have two updates expected in the same time as $i$'s first expected update. Recall that staleness corresponds to the difference between the version number of the current global model and the version number of the original global model that an agent's update was reported from. We derive the following expression for expected staleness of agent $i$.
\begin{gather}
    \mathbb{E}[\text{staleness of agent $i$ with rate $\tilde{r_i}$}] = \mathbb{E}[\tau_i] = \frac{\sum_{j=1}^n{\frac{\tilde{r_j}}{\tilde{r_i}}} - 1 }{b}
\end{gather}
Reconciling this with our mechanism definitions above, we get the unweighted influence of agent $i$ given its expected staleness.
\begin{gather}
    {p_i(\tilde{r_i})} = \frac{1}{\mathbb{E}[\tau_i] \cdot b + 1} 
\end{gather}
Recalling our derived weighting from the mechanism, we get a formula for our weighting as a function of expected staleness. We note that under a synchronous setting of zero expected staleness, we recover the exactly equal weighting used in synchronous FedAvg.
\begin{gather}
    \alpha(p_i(\tilde{r_i})) = \frac{\mathbb{E}[\tau_i] \cdot b + 1}{n}
\end{gather}

\subsubsection{Extending To Online Learning} \label{sec:online-learning}
We reconcile the derived weighting above into the online setting of arriving client updates. As noted in Section \ref{sec:step-size}, at any given aggregation, we should avoid updating the global model with a step size larger than any individual clients; otherwise, global model training can quickly diverge. As such, given a buffer of updates to be aggregated, we compute the desired weightings per the above and normalize such that they sum to 1.

However, this introduces complexity into our derivation from above. Intuitively, if an agent chooses to increase their expected staleness by slowing down their reporting speed, the best they can do is enjoy a single aggregation step to themselves of full step size, but consequently, they participate in fewer aggregations later due to their slower reporting speed. As a result, they contribute less to learning. We show this analytically: we can compute the normalized re-weighting for agent $i$'s update $\alpha_{\text{norm}}(p_i(\tilde{r}_{a_i}))$ in a buffer as the following:
\begin{align*}
\alpha_{\text{norm}}(p_i(\tilde{r}_{a_i})) 
    &=
    \frac{\mathbb{E}[\tau_{a_i}] \cdot b + 1}{\sum_{j=1}^b{\left[
    \mathbb{E}[\tau_{a_i}] \cdot b + 1
    \right]}} 
    = \frac{
        \left[\sum_{k=1}^n{
        \frac{\tilde{r}_k}{\tilde{r}_{a_i}} - 1
        }\right] + 1
    }{\sum_{j=1}^b{\left[
        \left[\sum_{k=1}^n{
        \frac{\tilde{r}_k}{\tilde{r}_{a_j}} - 1
        }\right] \
    + 1\right]}} \\
    &= \frac{
    \frac{1}{\tilde{r}_{a_i}}\sum_{k\neq a_i}^n{\tilde{r}_k} + 1
    }{
    b + \sum_{j=1}^n{\sum_{k\neq a_j}^n{\frac{\tilde{r}_k}{\tilde{r}_{a_j}}}}
    } 
\end{align*}
Isolating $\tilde{r}_{a_i}$, we get the following:
\begin{gather*}
    \alpha_{\text{norm}}(p_i(\tilde{r}_{a_i})) 
    =
    \frac{
    \frac{1}{\tilde{r}_{a_i}}\sum_{k\neq a_i}^n{\tilde{r}_k} + 1
    }{
    b + M
    }
    \\
    \text{ where } 
    M = \frac{1}{\tilde{r}_{a_i}}\sum_{k=1, k\neq a_i}^n {\tilde{r}_k} + \tilde{r}_{a_i} \sum_{j=1, a_j \neq a_i} ^ b {\frac{1}{\tilde{r}_{a_j}}} + \sum_{j=1, a_j \neq a_i}^b{
        \left[\sum_{k=1, k \notin [a_j, a_i]}^n {\frac{\tilde{r}_k}{\tilde{r}_{a_j}}}
        \right]
    }
\end{gather*}

We further note that, as an agent, by manipulating $\tilde{r}_{a_i}$ downwards, we increase expected staleness but also decrease the frequency of an agent's update. Recall from Section \ref{sec:fairness} that the utility of an agent is the normalized re-weighting times the proportional frequency of an agent's update based on the agent's reported update speed.
\begin{gather}
    u_{a_i}(\tilde{r}_{a_i}) =  \alpha_{\text{norm}}(p_i(\tilde{r}_{a_i})) 
 \cdot \left( \frac{
        \tilde{r}_{a_i}
    }{
        \tilde{r}_{a_i} + \sum_{j=1, a_j \neq a_i}^n{r_{a_j}}
    }\right) 
\end{gather}
Denoting $A = \sum_{k=1, k \neq a_i}^n{\tilde{r}_k}$, $C = \sum_{j=1, a_j \neq a_i}^b{1/\tilde{r}_{a_j}}$ and $D = \sum_{j=1, a_j \neq a_i}^b{
        \left[\sum_{k=1, k \notin [a_j, a_i]}^n {\frac{\tilde{r}_k}{\tilde{r}_{a_j}}}
        \right]
    }$, we take the derivative of the utility with respect to an agent's reporting speed to get:
\begin{gather*}
    \frac{\partial u_{a_i}}{\partial \tilde{r}_{a_i}}
    = \frac{A - C \tilde{r}_{a_i}^2}{(C\tilde{r}_{a_i}^2 + (b + D) \tilde{r}_{a_i} + A)^2}
\end{gather*}
Since our algorithm reweights based on expected staleness, we can rescale reported update rates $\tilde{r}_{a_i}$ linearly by any positive constant without changing the result of expected staleness (or our reweighting) to make $C \tilde{r}_{a_i}^2$ as small as possible and $A$ as large as possible. Furthermore, since $A$ increases with larger $n$, we observe with large enough $n$ (a large enough client pool) that $\partial u_{a_i}/\partial \tilde{r}_{a_i} > 0$, indicating that agents want to truthfully report as high of an update speed as possible and our aggregation mechanism is strictly strategy-proof with respect to agent manipulation of $\tilde{r}_{a_i}$.

\subsection{Algorithm}

We present the following algorithm, FedStaleWeight, applying the above weighting to each aggregation. In practice, to compute the expected staleness for each agent $i$, $\mathbb{E}[\tau_i]$, we maintain a moving average of previous staleness for that agent. Our method computes a weighting to apply over the aggregation buffer, meaning it can be combined with methods like weighted secure aggregation \cite{guo2021secure} to protect client gradients.

\begin{figure}[htbp]
\centering
\begin{minipage}{0.51\linewidth}
    \begin{algorithm}[H]
    \caption{\texttt{FedStaleWeight-server}}
    \label{algo:server}
    \SetAlgoLined
    \DontPrintSemicolon
    \textbf{Input: } buffer size $b$, number of aggregation rounds $N$, global learning rate $\eta_g$ initial model $w_0$\\
    Initialize $n \leftarrow 0$, $B = \{\}$ \;
    \While{n < N}{
        Receive $(\Delta_i, k)$ from any client $i$ and add to buffer $B$\;
        \If{$|B| = b$}{
            \For{$j = 1:b$}{
                $\alpha_j = \frac{\mathbb{E}[\tau_j] + 1}{n}$\;
            }
            $\alpha = \text{Normalize}(\{\alpha_1, ..., \alpha_b\})$ \;
            $w_{n+1} = w_{n} + \eta_g\sum_{b_i = (\Delta_j, \tau_j) \in B}{\alpha_i \Delta_j}$\;
            $B = \{\}, n \leftarrow n  + 1$\;
        }
    }
    \textbf{Output: } global model $w_N$\\
    \end{algorithm}
\end{minipage}
\hfill
\begin{minipage}{0.47\linewidth}
    \begin{algorithm}[H]
    \caption{\texttt{FedStaleWeight-client}}
    \label{algo:client}
    \SetAlgoLined
    \DontPrintSemicolon
    \textbf{Input: } client learning rate $\eta_\ell$, number of client SGD steps $Q$\\
    \While{Server Running}{
        Request latest model $w_k$ from server\;
        $y_0 \leftarrow w_k$\;
        \For{$q = 1:Q$}{
            $y_q = y_{q-1} - \eta_\ell g_{q}(y_{q-1})$
        }
        $\Delta = y_q - y_0$ \;
        Send $(\Delta, k)$ to server buffer \;
    }
    \end{algorithm}
\end{minipage}
\end{figure}

\section{Theoretical Results}

In this section, we provide a convergence guarantee for FedStaleWeight in the non-convex, smooth setting. Since FedStaleWeight effectively upweights stale updates in addition to operating in the asynchronous setting, it is essential to understand the relationship between convergence and our methods of reweighting to create asynchronous fairness. The full proof can be found in Appendix \ref{appendix:proof-of-convergence}.

\textbf{Notation. } We use the following notation and assumptions throughout as used in FedBuff \cite{nguyen2022federated}: $[n]$ represents the set of all clients, $\nabla F_i(w)$ is the gradient with respect to the loss on client $i$'s data. $f(w^*)$ is the minimum of $f(w)$, $g_i(w;\zeta_i)$ denotes the stochastic gradient on client $i$, $b$ is the buffer size for aggregation as noted above, and $Q$ is the number of local steps taken by each client. Similar to other seminal works in federated learning,  \cite{li2020federated} \cite{reddi2020adaptive} \cite{karimireddy2021scaffold} \cite{yu2018parallel}, we take the following assumptions: 

\begin{assumption}
    \label{assumption:unbiasedness}
    (Unbiasedness of client stochastic gradient) \; $\mathbb{E}_{\zeta}[g_i(w;\zeta_i)] = \nabla F_i(w)$
\end{assumption}

\begin{assumption}
    \label{assumption:bounded-local-global-variance}
    (Bounded local and global variance) for all clients $i \in [n]$, \; 
    \begin{gather*}
        \mathbb{E}_{\zeta_i}[\|g_i(w;\zeta_i) - \nabla F_i(w) \|^2] \leq \sigma_\ell^2
        \\
        \frac{1}{m}\sum_{i=1}^m\|\nabla F_i(w) - \nabla f(w)\|^2 \leq \sigma_g^2
    \end{gather*}
\end{assumption}

\begin{assumption}
    \label{assumption:bounded-gradient}
    (Bounded gradient) \; $\|\nabla F_i \|^2 \leq G$ \emph{ for all } $i \in [n]$
\end{assumption}

\begin{assumption}
    \label{assumption:lipschitz-gradient}
    (Lipschitz gradient) \; \emph{for all client } $i \in [n]$ \emph{ the gradient is L-smooth },
    \begin{gather*}
        \|\nabla F_i(w) - \nabla F_i(w') \|^2 \leq L \|w - w'\|^2
    \end{gather*}
\end{assumption}

\begin{assumption}
    \label{assumption:bounded-staleness}
    (Bounded Staleness) \; For all clients $i \in [n]$ and for each server step $t$, the staleness $\tau_i(t)$ between the model version a FedStaleWeight client uses to start local training and the model version in which the aggregated update $\Delta^i$ is used to modify the global version is not larger than $\tau_{\text{max}, 1}$ when $K=1$. Moreover, any buffered asynchronous aggregation with $b > 1$ has the maximum delay $\tau_{\text{max}, b}$ at most $\lceil \tau_{\text{max}, 1} / b \rceil$ \cite{nguyen2022federated}.
\end{assumption}

\begin{theorem}
    \label{theorem:convergence}
    Let $\eta_{\ell}^{(q)}$ be the local learning rate of client SGD in the q-th local step, and let $\alpha(Q) = \sum_{q=0}^{Q-1}{\eta_\ell^{(q)}}$, $\beta(Q) = \sum_{q=0}^{Q-1}{(\eta_\ell^{(q)})^2}$, and $U(\tau_{\max, b})$ be the maximum squared deviation between equal weighting ($1/b$) and a normalized $\alpha_j$ \ref{definition:umax}. Choosing $\eta_g \eta_\ell^{(q)} Q \leq 1/L$ for all local steps $q= 0, ..., Q-1$, the model iterates in Algorithm \ref{algo:server} achieves the following ergodic convergence rate:
    \begin{align*}
    \begin{split}
        \frac{1}{T}\sum_{t=0}^{T}{ \| \nabla f(w^t) \|^2}
    &\leq
    \frac{2 (f(w^0) - f(w^*))}{\eta_g \alpha(Q) T}
    \\
    &\qquad + 
    12 \eta_g^2 \tau_{\max, b}^2  \beta(Q) Q  L^2 
    \left( \frac{b\tau_{\max, b} + 1}{\tau_{\max, b} + 1}\right)^2
    \left(\sigma_\ell^2 + \sigma^2_g + G \right)
    \\
    &\qquad +
    12 \beta(Q)  Q L^2 ( \sigma_\ell^2 + \sigma_g^2 + G) 
    +
    4 b^2 U(\tau_{\max, b})G
    \end{split}
    \end{align*}
\end{theorem}

\section{Empirical Results}

\subsection{Simulation Details}
We implement a generic testbench to empirically verify our algorithm above. The simulation environment consists of concurrently running clients with private data and and a centralized server and buffer, where an aggregation strategy can be specified. 
\begin{enumerate}[leftmargin=2em]
    \item \textbf{Clients} run each on their own process thread concurrently, continuously pulling the latest available model from the global server and training it on their local private data before communicating the update to the server and repeating. Clients are initialized with both an individual slice of the overall dataset being evaluated (split into training and evaluation) and as a private runtime model representing a training delay distribution, from which the client process samples and sleeps for that duration to simulate local training delays.
    \item \textbf{Server} aggregates updates from the global buffer and updates the global model continuously. The server is initialized with a specified aggregation strategy and an aggregation buffer to pull updates from: once the buffer fills to a specific size, aggregation is performed and the global model is updated, after which clients will begin to pull that new model once they finish training. Clients broadcast their updates once training and their simulated delays complete, and their update is appended to the global buffer.
\end{enumerate}
The testbench allows users to specify custom aggregation strategies by implementing a simple callback function, from which evaluation against other baselines can be computed, logged and compared. We also implement the FedAvg baseline in this testbench.

\subsection{Dataset Distribution}

\begin{figure}[htbp]
    \centering
    \includegraphics[width=0.49\textwidth]{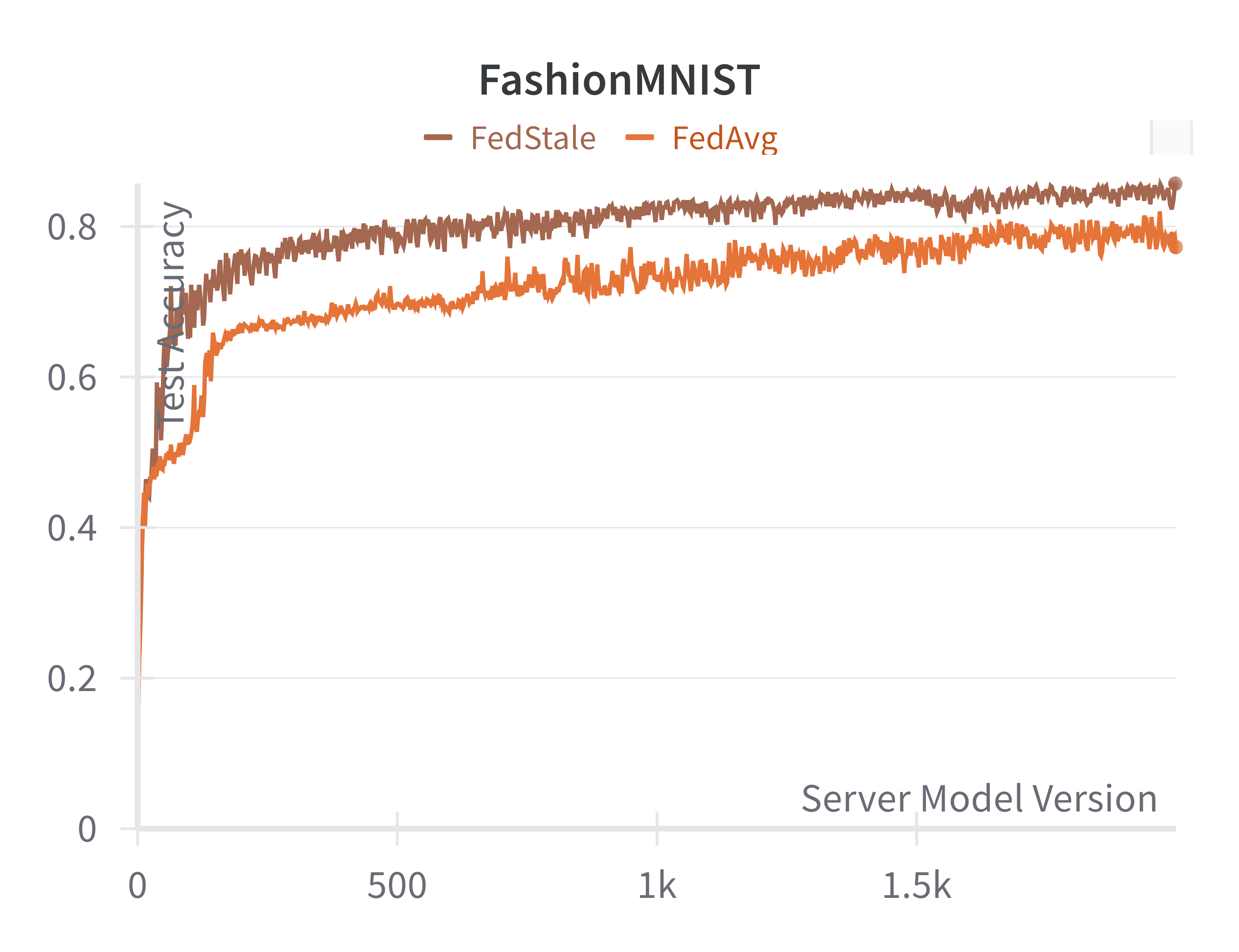}
    \includegraphics[width=0.49\textwidth]{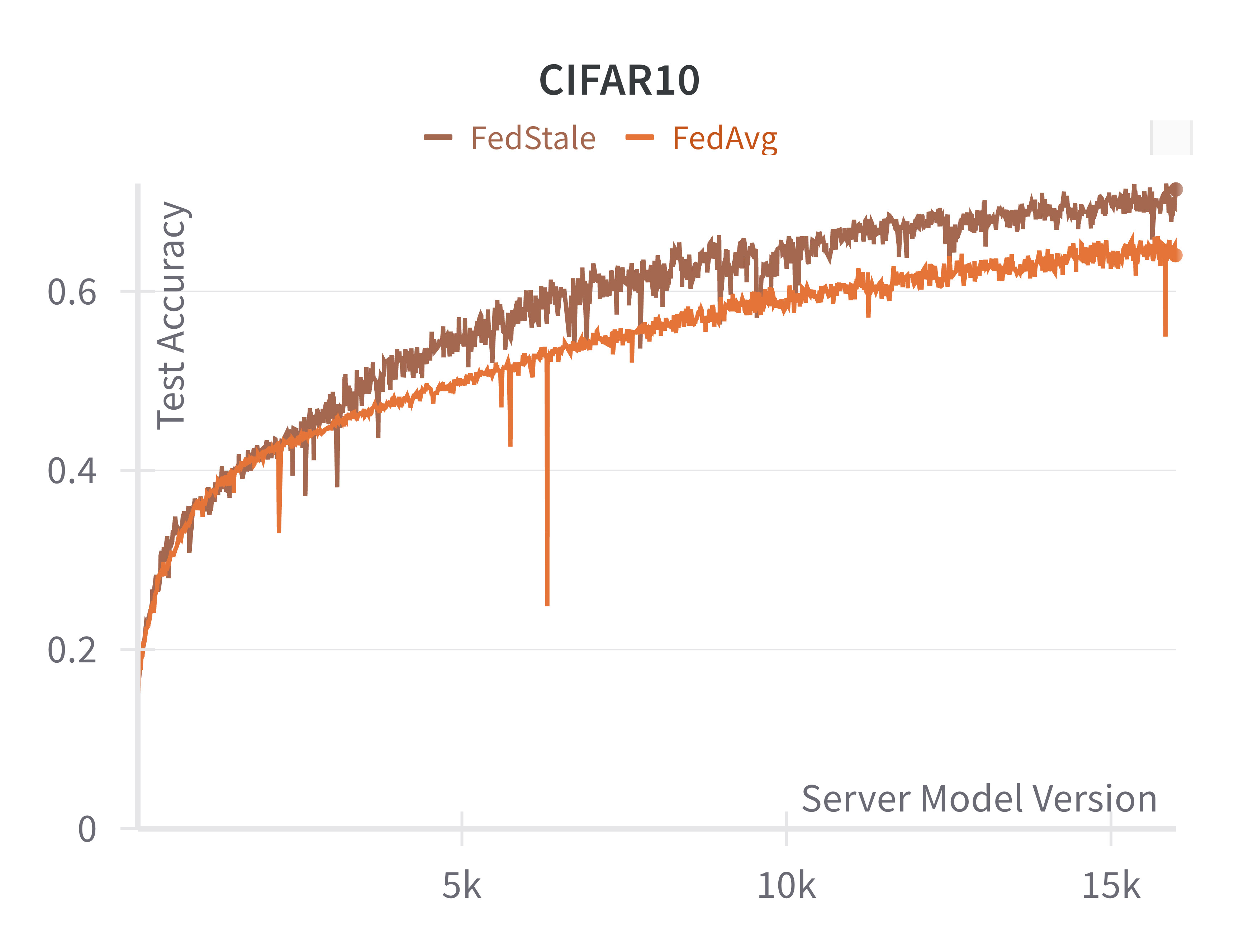}
    \caption{Example test accuracy curves for FedStaleWeight versus buffered FedAvg. FedStaleWeight more quickly converges to higher accuracy under the same non-IID setting, indicating higher fairness aggregating agent updates.}
    \label{fig:empirical_results}
\end{figure}

We test our algorithm empirically on two datasets, CIFAR10 and FashionMNIST. For both datasets, we model a system of 15 agents, with 10 ``fast" agents with training delay randomly sampled from uniform distributions $D_{\text{fast}} = U(1,2)$ and 5 ``slow" agents with training delays $D_{\text{slow}} = U(8,12)$. Data is distributed in a non-IID fashion and we reserve 20\% of data across all labels for testing global model accuracy to measure overall generalization For the remainder, the fast agents are given data points with labels 4 through 9, IID-distributed across fast agents where no two agents have the same data point. The slow agents are given labels 0 through 3, similarly IID-distributed. The results are shown in Figure \ref{fig:empirical_results}. 

Specifically, we model a scenario where low-compute agents have exclusive access to a non-trivial subset of all data: in this setting, higher fairness equates to a model with higher global accuracy, demonstrated in our empirical results. We run FashionMNIST and CIFAR10 for 4,000 and 16,000 aggregations respectively. We evaluate FedAvg and our FedStaleWeight with a buffer size of $b=5$ and a local learning rate of $\eta_\ell = 0.01$ with $Q=1$ local step. We clearly observe that our algorithm FedStaleWeight converges more quickly to a higher server test accuracy, indicating better generalization and less bias towards fast update producing agents.




\section{Discussion}

In this paper, we tackle the FL problem of training a model through updates received from a large number of edge devices. Our novel asynchronous technique for aggregating updates, FedStaleWeight, takes into account the staleness of client updates in order to re-weight them fairly. We argue that a slower client that sends fewer updates is undervalued in the aggregate under traditional averaging schemes, resulting in worse global models if the client's data set is unique; similarly, a faster client should not dominate the aggregate off of speed of compute alone. The algorithm utilizes a buffer that holds updates until full, at which time it performs the weighted aggregation computed by approximating expected staleness via a moving average and updates the global model accordingly.

By sending updates to the server, each client implicitly reports an ``update speed'' to a mechanism which is interpreted only via the update's staleness. This idea allows us to derive a re-weighting that increases fairness while still incentivizes truthfulness, ensuring that no agent wishes to intentionally slow down. This is critical to include when aggregation criteria uses the expected staleness as weighting, a inverse proxy for update speed. We show that the optimal weights are related to each client's expected staleness, a quantity easily estimable after multiple rounds have elapsed. We provide work towards verifying that truthfulness guarantees hold when extending to online learning and per buffer normalization by calibrating the mechanism parameters. 

Finally, we provide an ergodic convergence guarantee for FedStaleWeight and verify its efficacy through simulation. Our testbench simulates concurrent client threads and a server that executes our buffered aggregation algorithm while supporting the implementation of other aggregation techniques. This environment further serves as a useful resource for future work in the AFL space. In comparison to classic aggregation methods like FedAvg, we observe that FedStaleWeight converges faster and to a more accurate result, avoiding the pitfall of heavily weighting fast clients who may not be representative of the whole population.

There are several open questions, which we leave as future work to build upon. First, our framework assumes continuous client participation, when many real-world settings observe clients participating and dropping out frequently, necessitating more exploration into a scheme robust to probabilistic agent participation while maintaining fairness guarantees. We also assume that manipulation is zero-cost, when throttling could bring benefits of lower power consumption, affecting client utility. However, our work represents a strong step in narrowing the compute fairness gap between synchronous and asynchronous federated learning.

\section{Acknowledgements}
We are hugely grateful also to Safwan Hossain and the CS236R course for their support and mentorship during this project. This work was also supported by the FASRC computing cluster at Harvard University.

\bibliographystyle{unsrt} 
\nocite{*}
\bibliography{references}

\newpage

\appendix

\section*{Appendix}


\section{Proof of Convergence} \label{appendix:proof-of-convergence}
In this section, we prove the main convergence result for FedStaleWeight. Recall that FedStaleWeight updates can be described as the following update rule:
\begin{align*}
    w^{t+1} &= w^t + \eta_g \overline{\Delta}^t \\
    &= w_t + \eta_g \sum_{k \in S^t}\left( \alpha^{\text{norm}}_{k, t} \Delta_k^{t - \tau_k(t)}\right)
    \\
    &= w_t - \eta_g \sum_{k \in S^t}\left( \alpha^{\text{norm}}_{k, t} \sum_{q=0}^{Q-1} {\eta_\ell^{(q)} g_k \left(y_{k, q}^{t-\tau_k(t)}\right)}\right)
\end{align*}

\begin{table}
  \caption{Summary of notation}
  \label{sample-table}
  \centering
  \begin{tabular}{lll}
    \toprule                
    Description     & Symbol \\
    \midrule
    Number of server updates, server update index &  $T, t$  \\
   Set of clients contributing to update index $t$ &  $B_t$  \\
   Number of clients, client index & $n$, $i$ or $k$ \\
   Number of local steps per round, local step index & $Q, q$ \\
   Server model after $t$ steps & $w_t$ \\
   Stochastic gradient at client $i$ & $g_i (w; \zeta_i) := g_i(w)$ \\
   Local learning rate at local step $q$ & $\eta_{\ell}^{(q)}$ \\
   Global learning rate & $\eta_g$ \\
   Number of clients per update & b \\
   Local and global gradient variance & $\sigma_\ell^2, \sigma_g^2$ \\
   Delay or staleness of the client $i$'s model update for the $t$-th server update & $\tau_i(t)$ \\
   Maximum staleness for buffer size of & $\tau_{\max, b}$ \\
   Normalized FedStaleWeight re-weighting with respect to other elements \\ for buffer $B_t $ for agent $k$ such that $\sum_{k \in B_t}{\alpha^{\text{norm}}_{k, t}} = 1$ & $\alpha^{\text{norm}}_{k, t}$ \\
    \bottomrule
  \end{tabular}
\end{table}

As in \cite{nguyen2022federated}, in addition to our assumptions above, we assume that $B_t$ is a uniform subset of $[n]$: in other words, any client is equally likely to contribute in any given round. We can ensure this is the case in practice by, if a client contributes to a current update round, only allowing the server to sample that client once the current update is complete.

\begin{theorem}
    Let $\eta_{\ell}^{(q)}$ be the local learning rate of client SGD in the q-th local step, and let $\alpha(Q) = \sum_{q=0}^{Q-1}{\eta_\ell^{(q)}}$, $\beta(Q) = \sum_{q=0}^{Q-1}{(\eta_\ell^{(q)})^2}$, and $U(\tau_{\max, b})$ be the maximum squared deviation between equal weighting ($1/b$) and a normalized $\alpha_j$ \ref{definition:umax}. Choosing $\eta_g \eta_\ell^{(q)} Q \leq 1/L$ for all local steps $q= 0, ..., Q-1$, the model iterates in Algorithm \ref{algo:server} achieves the following ergodic convergence rate:
    \begin{align}
    \begin{split}
        \frac{1}{T}\sum_{t=0}^{T}{ \| \nabla f(w^t) \|^2}
    &\leq
    \frac{2 (f(w^0) - f(w^*))}{\eta_g \alpha(Q) T}
    \\
    &\qquad + 
    12 \eta_g^2 \tau_{\max, b}^2  \beta(Q) Q  L^2 
    \left( \frac{b\tau_{\max, b} + 1}{\tau_{\max, b} + 1}\right)^2
    \left(\sigma_\ell^2 + \sigma^2_g + G \right)
    \\
    &\qquad +
    12 \beta(Q)  Q L^2 ( \sigma_\ell^2 + \sigma_g^2 + G) 
    +
    4 b^2 U(\tau_{\max, b})G
    \end{split}
    \end{align}
\end{theorem}

We first restate a useful lemma from \cite{nguyen2022federated} which we use below.

\begin{lemma}
    $\mathbb{E}\left[\|g_k\|^2\right] \leq (\sigma_\ell^2 + \sigma^2_g + G)$, where the total expectation $\mathbb{E}[\cdot]$ is evaluated over the randomness with respect to client participation and the stochastic gradient taken by a client.
    \label{lemma:grad-norm-bound}
\end{lemma}

\emph{Proof.} By $L$-smoothness assumption,
\begin{align*}
    f(w^{t+1}) &\leq 
    f(w^{t}) + \eta_g \langle \nabla f(w^t), \overline{\Delta}^t \rangle + \frac{L \eta_g^2}{2} \left\|
        \overline{\Delta}^t 
    \right\|^2 \\
    &\leq f(w^{t}) 
    \underbrace{
        + \eta_g \sum_{k \in B_t}{\alpha^{\text{norm}}_{k, t} \left\langle 
            \nabla f(w^t), \Delta_k^{t-\tau_k}
        \right\rangle}
    }_{T_1} + 
    \underbrace{
        \frac{L \eta^2_g}{2} \left\| \sum_{k \in B_t}{\alpha^{\text{norm}}_{k, t} \Delta_k^{t - \tau_k}}\right\|^2
    }_{T_2}
\end{align*}

We then derive the upper bounds on $T_1$ and $T_2$. We expand $T_1$ as follows:
\begin{gather*}
    T_1 = \eta_g \sum_{k \in B_t}{\alpha^{\text{norm}}_{k, t} \left\langle 
            \nabla f(w^t), \Delta_k^{t-\tau_k}
        \right\rangle}
    = - \eta_g \sum_{k \in B_t}{
        \sum_{q=0}^{Q-1}{
            \alpha^{\text{norm}}_{k, t} \eta_\ell^{(q)}
            \left\langle 
                \nabla f(w^t), g_k\left(y_{k,q}^{t - \tau_k}\right)
            \right\rangle
        }
    }
\end{gather*}
Using conditional expectation, we can expand the expectation as
\begin{gather*}
    \mathbb{E}[\cdot] = \mathbb{E}_\mathcal{H}\mathbb{E}_{i \sim [n], g_i | i, \mathcal{H}}[\cdot]
\end{gather*}
where $\mathbb{E}_\mathcal{H}$ takes the expectation of the history of time-steps, $\mathbb{E}_{B_t\sim [n]}$ takes the expectation of the distribution of clients contributing at time-step $t$ over all $n$ clients and over the stochastic gradient of one step on a client.
\begin{align*}
    \mathbb{E}[T_1] 
    &= - \mathbb{E}\left[
        \eta_g \sum_{k \in B_t}{
        \sum_{q=0}^{Q-1}{
            \alpha^{\text{norm}}_{k, t} \eta_\ell^{(q)}
            \left\langle 
                \nabla f(w^t) ,\;
                g_k\left(y_{k,q}^{t - \tau_k}\right)
            \right\rangle
        }}
    \right] 
    \\
    &= - \eta_g \mathbb{E}_{\mathcal{H}, B_t \sim [n]}\left[
        \sum_{k\in B_t}{
        \alpha^{\text{norm}}_{i, t}
        \left[
            \sum_{q=0}^{Q-1}{
                 \eta_\ell^{(q)}
                 \mathbb{E}_{g_k | k, \mathcal{H}}
                \left\langle 
                    \nabla f(w^t) , \; 
                    g_k\left(y_{k,q}^{t - \tau_k}\right)
                \right\rangle
            }
        \right]
        }
    \right]
    \\
    &= - \eta_g \mathbb{E}_{\mathcal{H}, B_t \sim [n]}\left[ 
        \sum_{q=0}^{Q-1}{
                \eta_\ell^{(q)}
                \left\langle 
                    \nabla f(w^t), \; 
                    \alpha^{\text{norm}}_{k, t} \sum_{k \in B_t}{
                    \nabla F_k\left(y_{k, q}^{t- \tau_k}\right)}
                \right\rangle
        }
    \right]
\end{align*}
Using the identity
\begin{gather*}
    \langle a, b\rangle = \frac{1}{2}\left(
        \|a\|^2 + \|b\|^2 - \|a - b\|^2
    \right)
\end{gather*}
we further expand as follows:
\begin{align*}
    \mathbb{E}[T_1] 
    &= -\frac{\eta_g}{2}\left(
        \sum_{q=0}^{Q-1}{\eta_\ell^{(q)}} \right) {\| \nabla f(w^t) \|^2}  
    \underbrace{- \sum_{q=0}^{Q-1}{
            \frac{
                \eta_g \eta_\ell^{(q)}
            }{
                2            
            }\bigg(
                \mathbb{E}_{\mathcal{H}, B_t \sim [n]}
                \left\|
                    \sum_{k \in B_t}{\alpha^{\text{norm}}_{k, t} \; \nabla F_k\left(y_{k, q}^{t - \tau_k}\right)}
                \right\|^2
        } \bigg)}_{T_3}
    \\
    &+
    \sum_{q=0}^{Q-1}{
            \frac{
                \eta_g \eta_\ell^{(q)}
            }{
                2            
            }}\bigg(
    \mathbb{E}_{\mathcal{H}, B_t \sim [n]} 
    \left\|
        \nabla f(w^t) - \sum_{k \in B_t}{
            \alpha^{\text{norm}}_{k, t} \; \nabla F_k(y_{k, q}^{t - \tau_k})
        }
    \right\|^2
    \bigg)
\end{align*}
We expand $T_2$ as follows:
\begin{align*}
    \mathbb{E}[T_2]
    &= \mathbb{E}\left[
    \frac{L \eta^2_g}{2} \left\| \sum_{k \in B_t}{\alpha^{\text{norm}}_{k, t} \Delta_k^{t - \tau_k}}\right\|^2
    \right]
    \\
    &= 
    \frac{L \eta^2_g}{2} \cdot 
    \mathbb{E}_{\mathcal{H}, B_t \sim [n]}
    \left\|
        \sum_{k\in B_t}{
        \alpha_{k, t}^\text{norm} \left(
            \sum_{q=0}^{Q-1}{
                -\eta_\ell^{(q)} \mathbb{E}_{g_k | \mathcal{H}, B_t}[g_k(y_{k, q}^{t - \tau_k})]
            }
        \right)
        }
    \right\|^2
    \\
    &= 
    \frac{L \eta^2_g}{2} \cdot 
    \mathbb{E}_{\mathcal{H}, B_t \sim [n]}
    \left\|
        \sum_{q=0}^{Q-1}{
        \eta_\ell^{(q)} 
        \sum_{k\in B_t}{
        \alpha_{k, t}^\text{norm} \left(
                \nabla F_k(y_{k, q}^{t - \tau_k})
        \right)
        }
        }
    \right\|^2
    \\
    &\leq
    \sum_{q=0}^{Q-1}{\frac{QL \eta^2_g \left(\eta_\ell^{(q)}\right)^2 }{2} \cdot 
    \mathbb{E}_{\mathcal{H}, B_t \sim [n]}
        \left\|
        \sum_{k\in B_t}{
        \alpha_{k, t}^\text{norm} \left(
                \nabla F_k(y_{k, q}^{t - \tau_k})
        \right)
        }
    \right\|^2}
\end{align*}

We can choose step sizes such that $\mathbb{E}[T_2] + \mathbb{E}[T_3] \leq 0$, meaning that we can choose $\eta_g \eta_\ell^{(q)} Q \leq 1 / L$ for all $q \in [0, ..., Q-1]$. We then get the following inequality:
\begin{align*}
    f(w^{t+1})
    &\leq f(w^{t}) 
    -\frac{\eta_g}{2}\left(
        \sum_{q=0}^{Q-1}{\eta_\ell^{(q)}} \right) {\| \nabla f(w^t) \|^2} 
    \\    
    &\qquad + 
    \sum_{q=0}^{Q-1}{
            \frac{
                \eta_g \eta_\ell^{(q)}
            }{
                2            
            }}\bigg(
    \mathbb{E}_{\mathcal{H}, B_t \sim [n]} 
    \underbrace{\left\|
        \nabla f(w^t) - \sum_{k \in B_t}{
            \alpha^{\text{norm}}_{k, t} \; \nabla F_k(y_{k, q}^{t - \tau_k})
        }
    \right\|^2}_{T_4}
    \bigg)
\end{align*}
We can telescope and expand $T_4$ as follows. Note that this variance term after telescoping breaks down into four key portions:
\allowdisplaybreaks
\begin{align*}
    \mathbb{E}_{\mathcal{H}, B_t \sim [n]}[T_4]
    &= \mathbb{E}_{\mathcal{H}, B_t \sim [n]}
        \left\|
        \nabla f(w^t) - \sum_{k \in B_t}{
            \alpha^{\text{norm}}_{k, t} \; \nabla F_k(y_{k, q}^{t - \tau_k})
        }
        \right\|^2
    \\
    &= \mathbb{E}_{\mathcal{H}, B_t \sim [n]}
        \Bigg\|
        \frac{1}{n}\sum_{i=1}^n \nabla F_i(w^t)
        - 
        \frac{1}{n}\sum_{i=1}^n \nabla F_i(w^{t - \tau_i})
        +
        \frac{1}{n}\sum_{i=1}^n \nabla F_i(w^{t - \tau_i})
    \\
        &\qquad\qquad 
        -
        \frac{1}{|B_t|} \sum_{k \in B_t}{\nabla F_k(w^{t - \tau_k})}
        +
        \frac{1}{|B_t|} \sum_{k \in B_t}{\nabla F_k(w^{t - \tau_k})}
    \\
        &\qquad\qquad 
        -
        \frac{1}{|B_t|} \sum_{k \in B_t}{\nabla F_k(y^{t - \tau_k}_{k, q}})
        +
        \frac{1}{|B_t|} \sum_{k \in B_t}{\nabla F_k(y^{t - \tau_k}_{k, q}})
        -
        \sum_{k \in B_t}{
            \alpha^{\text{norm}}_{k, t} \; \nabla F_k(y_{k, q}^{t - \tau_k})
        }
        \Bigg\|^2
    \\
    &\leq 4 \Biggl(
    \underbrace{\mathbb{E}_{\mathcal{H}, B_t \sim [n]}
        \Bigg\|
        \frac{1}{n}\sum_{i=1}^n \nabla F_i(w^t)
        - 
        \frac{1}{n}\sum_{i=1}^n \nabla F_i(w^{t - \tau_i})
        \Bigg\|^2}_{\text{staleness error}} 
    \\
        &\qquad\qquad +
        \underbrace{\mathbb{E}_{\mathcal{H}, B_t \sim [n]}
        \Bigg\|
        \frac{1}{n}\sum_{i=1}^n \nabla F_i(w^{t - \tau_i})
        -
        \frac{1}{|B_t|} \sum_{k \in B_t}{\nabla F_k(w^{t - \tau_k})}
        \Bigg\|^2}_{\text{sampling error}}  
    \\
        &\qquad\qquad +
        \underbrace{\mathbb{E}_{\mathcal{H}, B_t \sim [n]}
        \Bigg\|
        \frac{1}{|B_t|} \sum_{k \in B_t}{\nabla F_k(w^{t - \tau_k})}
        -
        \frac{1}{|B_t|} \sum_{k \in B_t}{
             \nabla F_k(y_{k, q}^{t - \tau_k})
        }
        \Bigg\|^2}_{\text{local client drift}} 
    \\
        &\qquad\qquad +
        \underbrace{\mathbb{E}_{\mathcal{H}, B_t \sim [n]}
        \Bigg\|
        \frac{1}{|B_t|} \sum_{k \in B_t}{
             \nabla F_k(y_{k, q}^{t - \tau_k})
        }
        -
        \sum_{k \in B_t}{
            \alpha^{\text{norm}}_{k, t} \; \nabla F_k(y_{k, q}^{t - \tau_k})
        }
        \Bigg\|^2}_{\text{reweighting error}}
        \Biggl)
\end{align*}
Using our assumptions on $L$-smoothness, we can further reduce this as follows. We note that due to unbiasedness of the sample mean, on average, the mean over all agent's gradients is equal to the mean of the gradients of agents sampled at each stage.
\begin{align*}
    \mathbb{E}_{\mathcal{H}, B_t\sim [n]}[T_4]
    &\leq
    \frac{4L^2}{n}\sum_{i=1}^n \mathbb{E}_{\mathcal{H}}\left\|
    w^t - w^{t-\tau_i}
    \right\|^2 +
        \frac{4L^2}{b}
        \sum_{k \in B_t}
        {
        \mathbb{E}_{\mathcal{H}, B_t \sim [n]}
        \Bigg\| w^{t-\tau_k} - y_{k, q}^{t - \tau_k} \Bigg\|^2
        }      
    \\
        &\qquad\qquad +
        4 \mathbb{E}_{\mathcal{H}, B_t \sim [n]}
        \underbrace{
        \Bigg\| 
        \frac{1}{|B_t|} \sum_{k \in B_t}{
             \nabla F_k(y_{k, q}^{t - \tau_k})
        }
        -
        \sum_{k \in B_t}{
            \alpha^{\text{norm}}_{k, t} \; \nabla F_k(y_{k, q}^{t - \tau_k})
        }
        \Bigg\|^2
        }_{T_5}
\end{align*}

We can produce an upper-bound on the staleness error (first term above) as follows:
\begin{align*}
    \left\|
        w^t - w^{t - \tau_i}
    \right\|^2
    &= 
    \left\|
        \sum_{\rho = t - \tau_I}^{t-1}
        {
            (w^{\rho + 1} - w^\rho)
        }
    \right\|^2
    \\
    &=
    \eta_g^2 \left\|
        \sum_{\rho = t - \tau_i}^{t-1}
        {
           \sum_{j_\rho \in S_\rho}
           {
                \alpha_{j_\rho, \rho}^\text{norm}
                \sum_{l=0}^{Q-1}
                {
                \eta_\ell^{(l)}
                g_{j_\rho}(y^\rho_{j_\rho, l})
                }
           }
        }
    \right\|^2
\end{align*}
Taking the expectation of this with respect to $\mathcal{H}$, we get:
\begin{align*}
    \mathbb{E}_\mathcal{H}\left\|
        w^t - w^{t - \tau_i}
    \right\|^2
    &\leq 
    \eta_g^2 b \tau_i 
        \sum_{\rho = t - \tau_i}^{t-1}
        {
           \sum_{j_\rho \in S_\rho}
           {
                (\alpha_{j_\rho, \rho}^\text{norm})^2
                \cdot
                \mathbb{E}_\mathcal{H}
                \left\|\sum_{l=0}^{Q-1}
                {
                \eta_\ell^{(l)}
                g_{j_\rho}(y^\rho_{j_\rho, l})
                }
                 \right\|^2
           }
        }
\end{align*}
Recalling our definition of $\alpha^{\text{norm}}_{j_\rho, \rho}$ and our assumption on bounded staleness, we can see that given some max-staleness $\tau_{\max, b}$, we see that the largest that $\alpha^{\text{norm}}_{j_\rho, \rho}$ can be is an aggregation buffer with one agent with maximum expected staleness $\tau_{\max, b}$ with all other agents having zero staleness (and $1/n$ un-normalized weighting).
\begin{align*}
    \alpha^{\text{norm}}_{j_\rho, \rho} 
    &\leq \frac{
        \frac{\tau_{\max, b} \cdot b + 1}{n}
    }{
        \frac{\tau_{\max, b} \cdot b + 1}{n} + \sum_{i=1}^{b-1}{\frac{1}{m}}
    }
    \\
    &\leq \frac{b\tau_{\max, b} + 1}{b(\tau_{\max, b} + 1)}
\end{align*}
The expectation for the staleness error becomes bounded as follows, using the above result and Lemma \ref{lemma:grad-norm-bound}:
\begin{align*}
    \mathbb{E}_\mathcal{H}\left\|
        w^t - w^{t - \tau_i}
    \right\|^2
    &\leq 
    \eta_g^2 b \tau_i Q
        \sum_{\rho = t - \tau_i}^{t-1}
        {
           \sum_{j_\rho \in S_\rho}
           {
                \left( \frac{b\tau_{\max, b} + 1}{b(\tau_{\max, b} + 1)}\right)^2
                \sum_{l=0}^{Q-1}
                {
                \left(
                    \eta_\ell^{(l)}
                \right)^2
                \cdot
                \mathbb{E}_\mathcal{H}
                \left\|
                g_{j_\rho}(y^\rho_{j_\rho, l})
                 \right\|^2
                } 
           }
        }
    \\ 
    &\leq 
    \eta_g^2 b^2 \max_{\tau_i}{\tau_i^2} Q
            \left( \frac{b\tau_{\max, b} + 1}{b(\tau_{\max, b} + 1)}\right)^2
            \left(
            \sum_{l=0}^{Q-1}
            {
                \left(
                    \eta_\ell^{(l)}
                \right)^2
            }
            \right)
            3\left(\sigma_\ell^2 + \sigma^2_g + G \right)
    \\
    &\leq 
    3Q\eta_g^2 \tau_{\max, b}^2 
            \left( \frac{b\tau_{\max, b} + 1}{\tau_{\max, b} + 1}\right)^2
            \left(
            \sum_{l=0}^{Q-1}
            {
                \left(
                    \eta_\ell^{(l)}
                \right)^2
            }
            \right)
            \left(\sigma_\ell^2 + \sigma^2_g + G \right)
\end{align*}
Likewise for the local drift error, we can bound it similarly using Lemma \ref{lemma:grad-norm-bound}:
\begin{align*}
    \mathbb{E} \| w^{t-\tau_i} - y_{i, q}^{t - \tau_i} \|^2 &=  \mathbb{E} \| y_{i,0}^{t - \tau_i} - y_{i, q}^{t - \tau_i} \|^2 = \mathbb{E}\left\|
        \sum_{l=0}^{q-1}{
         \eta_\ell^{(l)} g_i(y_{i, l}^{t - \tau_i)}
        }
    \right\|^2 
    \\
    &\leq 
    3q \left(
        \sum_{i=0}^{q-1}{\left(\eta_\ell^{(i)}\right)^2}
    \right) ( \sigma_\ell^2 + \sigma_g^2 + G)
    \\
    &\leq 
    3Q \left(
        \sum_{i=0}^{Q-1}{\left(\eta_\ell^{(i)}\right)^2}
    \right) ( \sigma_\ell^2 + \sigma_g^2 + G)
\end{align*}
Finally, we can bound the re-weighting error as follows:
\begin{align*}
    \mathbb{E}[T_5]
    &= \mathbb{E} \left\|
         \sum_{k \in B_t}{
             \left( \frac{1}{|B_t|} -  \alpha^{\text{norm}}_{k, t}\right)\nabla F_k(y_{k, q}^{t - \tau_k})
        }
    \right\|^2
    \\
    &\leq \mathbb{E}\left[
        \sum_{k \in B_t}{
            \left\|
                \left(
                    \frac{1}{|B_t|} -  \alpha^{\text{norm}}_{k, t}
                \right)\nabla F_k(y_{k, q}^{t - \tau_k})
            \right\|^2
        }
    \right]
    \\
    &\leq b \cdot \mathbb{E}\left[
        \sum_{k \in B_t}{
            \left(
                \frac{1}{b} -  \alpha^{\text{norm}}_{k, t}
            \right)^2 
            \left\|
                \nabla F_k(y_{k, q}^{t - \tau_k})
            \right\|^2
        }
    \right]
\end{align*}
We note from before that the largest that $\alpha_{k, t}^{norm}$ can be is $\frac{b\tau_{\max, b} + 1}{b(\tau_{\max, b} + 1)}$. Consequently, the smallest that $\alpha_{k, t}^{norm}$ can be is an agent with zero expected staleness (minimizing the numerator), with all other buffer agents having maximum expected staleness (maximizing the denominator):
\begin{align*}
   \alpha_{k, t}^{norm} 
   &\geq 
   \frac{\frac{1}{n}}{\frac{1}{n} + \sum_{i=1}^{b-1}{\frac{b\tau_{\max, b} + 1}{n}}}
    \\
    &\geq \frac{1}{(b-1)(b\tau_{\max, b} + 1) + 1}
\end{align*}
We can then upper-bound as follows, denoting this as quantity $U(\tau_{\max, b})$:
\begin{gather}
    \left(\frac{1}{b} - \alpha_{k, t}^{\text{norm}}\right)^2 
    \leq \max\left\{
        \left(
            \frac{1}{b} - \frac{b\tau_{\max, b} + 1}{b(\tau_{\max, b} + 1)}
        \right)^2,
        \left(
            \frac{1}{b} - \frac{1}{(b-1)(b\tau_{\max, b} + 1) + 1}
        \right)^2
    \right\} = U(\tau_{\max, b})
    \label{definition:umax}
\end{gather}
We arrive at the following expectation of $T_5$, given the above result and Assumption \ref{assumption:bounded-gradient}:
\begin{gather*}
    \mathbb{E}[T_5]
    \leq U(\tau_{\max, b}) b^2  G
\end{gather*}
Thus, combining everything, denoting $\alpha(Q) = \sum_{q=0}^{Q-1}(\eta_{\ell}^{(q)})$ and $\beta(Q) = \sum_{q=0}^{Q-1}(\eta_{\ell}^{(q)})^2$, we have
\begin{align*}
    \mathbb{E}[f(w^{t+1})]
    &\leq 
    \mathbb{E}[f(w^{t})]
    -\frac{\eta_g}{2}\alpha(Q) { \mathbb{E}\| \nabla f(w^t) \|^2} 
    \\    
    &\qquad + 
    \sum_{q=0}^{Q-1}{
            \frac{
                \eta_g \eta_\ell^{(q)}
            }{
                2            
            }}\bigg(
    \frac{4L^2}{n}\sum_{i=1}^n \mathbb{E}_{\mathcal{H}}\left\|
    w^t - w^{t-\tau_i}
    \right\|^2 +
        \frac{4L^2}{b}
        \sum_{k \in B_t}
        {
        \mathbb{E}_{\mathcal{H}, B_t \sim [n]}
        \Bigg\| w^{t-\tau_k} - y_{k, q}^{t - \tau_k} \Bigg\|^2
        }      
    \\
        &\qquad\qquad +
        4 \mathbb{E}_{\mathcal{H}, B_t \sim [n]}
        \Bigg\| 
        \frac{1}{|B_t|} \sum_{k \in B_t}{
             \nabla F_k(y_{k, q}^{t - \tau_k})
        }
        -
        \sum_{k \in B_t}{
            \alpha^{\text{norm}}_{k, t} \; \nabla F_k(y_{k, q}^{t - \tau_k})
        }
        \Bigg\|^2  
    \bigg)
    \\
    &\leq 
    \mathbb{E}[f(w^{t})]
    -\frac{\eta_g}{2}\alpha(Q) {\mathbb{E}\| \nabla f(w^t) \|^2} 
    \\    
    &\qquad + 
    \sum_{q=0}^{Q-1}{
            \frac{
                \eta_g \eta_\ell^{(q)}
            }{
                2            
            }}
    \bigg(
    12L^2 
    Q\eta_g^2 \tau_{\max, b}^2  \beta(Q)
            \left( \frac{b\tau_{\max, b} + 1}{\tau_{\max, b} + 1}\right)^2
            \left(\sigma_\ell^2 + \sigma^2_g + G \right)
    \bigg)
    \\
    &\qquad +
    \sum_{q=0}^{Q-1}{
            \frac{
                \eta_g \eta_\ell^{(q)}
            }{
                2            
            }}
    \bigg(
        12QL^2
         \left(
            \sum_{i=0}^{Q-1}{\left(\eta_\ell^{(i)}\right)^2}
        \right) ( \sigma_\ell^2 + \sigma_g^2 + G)
    \bigg)
    \\
    &\qquad
     +
    \sum_{q=0}^{Q-1}{
            \frac{
                \eta_g \eta_\ell^{(q)}
            }{
                2            
            }}
    \bigg(
        4  U(\tau_{\max, b}) b^2  G
    \bigg)
    \\
    &\leq 
    \mathbb{E}[f(w^{t})] 
    -\frac{\eta_g}{2}\alpha(Q) {\mathbb{E}\| \nabla f(w^t) \|^2} 
    \\    
    &\qquad + 
    6 \eta_g^3 \tau_{\max, b}^2  \alpha(Q) \beta(Q) Q  L^2 
    \left( \frac{b\tau_{\max, b} + 1}{\tau_{\max, b} + 1}\right)^2
    \left(\sigma_\ell^2 + \sigma^2_g + G \right)
    \\
    &\qquad +
    6 \eta_g \alpha(Q) \beta(Q)  Q L^2 ( \sigma_\ell^2 + \sigma_g^2 + G) 
    \\
    &\qquad 
    +
     2 b^2 \eta_g  U(\tau_{\max, b}) \alpha(Q) G
\end{align*}
Rearranging and summing from $t=1$ to $T$, we get the following:
\begin{align*}
    \sum_{t=0}^{T}{\eta_g \alpha(Q) \| \nabla f(w^t) \|^2}
    &\leq
    \sum_{t=0}^{T-1} 2\left(
        \mathbb{E}[f(w^t) - \mathbb{E}[f(w^{t+1})]]
    \right)
    \\
    &\qquad + 
    12 T \eta_g^3 \tau_{\max, b}^2  \alpha(Q) \beta(Q) Q  L^2 
    \left( \frac{b\tau_{\max, b} + 1}{\tau_{\max, b} + 1}\right)^2
    \left(\sigma_\ell^2 + \sigma^2_g + G \right)
    \\
    &\qquad +
    12 T \eta_g \alpha(Q) \beta(Q)  Q L^2 ( \sigma_\ell^2 + \sigma_g^2 + G) 
    \\
    &\qquad 
    +
    4 T b^2 \eta_g  U(\tau_{\max, b}) \alpha(Q) G
\end{align*}
Thus, we conclude
\begin{align*}
    \frac{1}{T}\sum_{t=0}^{T}{ \| \nabla f(w^t) \|^2}
    &\leq
    \frac{2 (f(w^0) - f(w^*))}{\eta_g \alpha(Q) T}
    \\
    &\qquad + 
    12 \eta_g^2 \tau_{\max, b}^2  \beta(Q) Q  L^2 
    \left( \frac{b\tau_{\max, b} + 1}{\tau_{\max, b} + 1}\right)^2
    \left(\sigma_\ell^2 + \sigma^2_g + G \right)
    \\
    &\qquad +
    12 \beta(Q)  Q L^2 ( \sigma_\ell^2 + \sigma_g^2 + G) 
    +
    4 b^2 U(\tau_{\max, b})G
\end{align*}
\qed
\end{document}